\documentclass[conference]{IEEEtran}

\makeatletter
\def\ps@IEEEtitlepagestyle{%
  \def\@oddfoot{\mycopyrightnotice}%
  \def\@evenfoot{}%
}
\def\mycopyrightnotice{%
  {\footnotesize 979-8-3503-2297-2/23/\$31.00~\copyright~2023 IEEE\hfill}
  \gdef\mycopyrightnotice{}
}

\usepackage{blindtext}
\usepackage{eso-pic}
\IEEEoverridecommandlockouts
\usepackage{cite}
\usepackage{amsmath,amssymb,amsfonts}
\usepackage{algorithmic}
\usepackage{graphicx}
\usepackage{textcomp}
\usepackage{xcolor}
\usepackage{makecell}
\usepackage{subcaption}
\usepackage{multirow}
\usepackage{url} 
\def\BibTeX{{\rm B\kern-.05em{\sc i\kern-.025em b}\kern-.08em
    T\kern-.1667em\lower.7ex\hbox{E}\kern-.125emX}}
    
\usepackage{eso-pic}
\newcommand\AtPageUpperMyright[1]{\AtPageUpperLeft{%
 \put(\LenToUnit{0.17\paperwidth},\LenToUnit{-2cm}){%
     \parbox{0.9\textwidth}{\raggedleft\fontsize{8}{11}\selectfont #1}}%
 }}%
\newcommand{\conf}[1]{%
\AddToShipoutPictureBG*{%
\AtPageUpperMyright{#1}
}
}    

\newcommand{\linebreakand}{%
  \end{@IEEEauthorhalign}
  \hfill\mbox{}\par
  \mbox{}\hfill\begin{@IEEEauthorhalign}
}

\begin{document}
\title{\vspace*{1cm} Adversarial Attacks on Traffic Sign Recognition: \\A Survey}


\author{\IEEEauthorblockN{Svetlana Pavlitska}
\IEEEauthorblockA{
\textit{FZI Research  Center  for  Information  Technology}\\
\textit{Karlsruhe Institute of Technology (KIT)}\\
Karlsruhe, Germany\\
pavlitska@fzi.de}
\and
\IEEEauthorblockN{Nico Lambing}
\IEEEauthorblockA{
\textit{FZI Research  Center  for  Information  Technology}\\
Karlsruhe, Germany\\
lambing@fzi.de}
\linebreakand 
\IEEEauthorblockN{J. Marius Z\"ollner}
\IEEEauthorblockA{\textit{FZI Research  Center  for  Information  Technology}\\
\textit{Karlsruhe Institute of Technology (KIT)}\\
Karlsruhe, Germany \\
zoellner@fzi.de}
}

\maketitle
\thispagestyle{plain}
\pagestyle{plain}
\conf{\textit{  Proc. of the International Conference on Electrical, Computer, Communications and Mechatronics Engineering (ICECCME 2023) \\ 
19-20 July 2023, Tenerife, Canary Islands, Spain}}
\begin{abstract}
Traffic sign recognition is an essential component of perception in autonomous vehicles, which is currently performed almost exclusively with deep neural networks (DNNs). However, DNNs are known to be vulnerable to adversarial attacks. Several previous works have demonstrated the feasibility of adversarial attacks on traffic sign recognition models. Traffic signs are particularly promising for adversarial attack research due to the ease of performing real-world attacks using printed signs or stickers. In this work, we survey existing works performing either digital or real-world attacks on traffic sign detection and classification models. We provide an overview of the latest advancements and highlight the existing research areas that require further investigation.
\end{abstract}


\section{Introduction}
Deep neural networks (DNNs) are inherently susceptible to adversarial attacks: small changes in the input can cause wrong model predictions~\cite{szegedy2013intriguing,goodfellow2014explaining}. To obstruct the behavior of a DNN for computer vision tasks, an attacker can perform either small imperceptible pixel changes over the whole image, or restrict visible, perceptible adversarial noise to a small image area, thus generating an \textit{adversarial patch}~\cite{brown2017adversarial}. Real-world attacks can easily be performed by adding a printed patch to a scene perceived by a camera.

Correct traffic sign recognition is vital for perception in automated vehicles. Unlike many other perception tasks, traffic sign recognition can be performed only using input from camera images, and thus errors cannot be compensated by other sensors like LiDAR. 
Adversarial attacks on computer vision models thus pose a special threat to this perception task. On the other hand, attacks against traffic sign recognition are especially favorable regarding real-world evaluation since perturbed traffic signs can easily be printed to replace the real ones. Therefore, a particular research effort has been put into designing attacks that would be robust under various physical conditions.  




\begin{figure}[t]
\centering
 \resizebox{1.0\linewidth}{!}{
 \begin{tabular}{c c c}
    \includegraphics[width=0.2\textwidth]{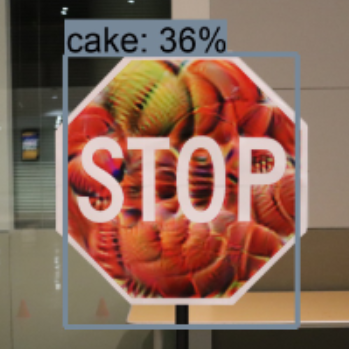} &
    \includegraphics[width=0.2\textwidth]{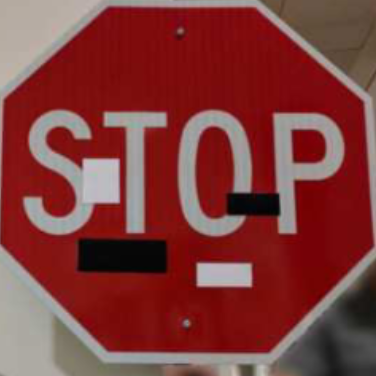} &
    \includegraphics[width=0.2\textwidth]{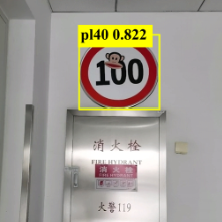} \\

  	a) Attack on the \textbf{generic}  & 
    b) Attack on the \textbf{traffic}   & 
    c) Attack on the \textbf{traffic} \\ 
    \textbf{object detection}: a stop & \textbf{sign classification}: a & \textbf{sign detection and} \\
    sign misclassified by the & stop sign misclassified  & \textbf{classification}: a 100km/h  \\
    Faster R-CNN trained  & as a speed limit 45km/h  & sign is misclassified as \\
    on the COCO dataset~\cite{chen2018shapeshifter}. & sign by LISA-CNN~\cite{eykholt2018robust}.&  a 40km/h  sign~\cite{wei2023adversarial}.\\
    \end{tabular}
    }
    \caption{Three subtasks within TSR and exemplary attacks.}
    \label{fig:groups}
\end{figure}


\textit{Traffic sign recognition (TSR)} is a multi-class classification problem with unbalanced class frequencies. TSR is split into two tasks: \textit{traffic sign detection (TSD)} aims at traffic sign localization in the input image, while the goal of the next step, \textit{traffic sign classification (TSC)} is to predict class labels for the detections (see Figure~\ref{fig:groups}). Traffic sign recognition is performed predominantly with DNNs, mostly convolutional neural networks (CNNs)~\cite{sermanet2011traffic}, and recently also spatial transformers~\cite{garcia2018deep}. Successful adversarial attacks were demonstrated both for traffic sign classification~\cite{eykholt2018robust} and detection~\cite{wei2023adversarial} models, while attacks took different forms: an attacker can perturb the whole area of a traffic sign~\cite{lu2017adversarial}, generate stickers~\cite{eykholt2018robust} or shadows\cite{zhong2022shadows} to be placed into it.    

This survey provides a comprehensive overview of the existing works on adversarial attacks on TSR models. We compare models, datasets, and attack approaches. Furthermore, we identify promising research directions. To the best of our knowledge, we are the first to provide such an overview.

\begin{table*}[t]
\centering
\caption{Overview of traffic sign recognition datasets}
\label{tab:datasets}
  \resizebox{1.0\linewidth}{!}{
\begin{tabular}{|l   c   c  |c  c   c  c   c   c| }
\hline
\textbf{Dataset} & \textbf{Year} & \textbf{Ref} & \textbf{TSD}  & \textbf{TSC} & \textbf{Images annotated} & \textbf{Classes}  & \textbf{Resolution} & \textbf{Country} \\ \hline

GTSRB & 2011 &~\cite{stallkamp2011gtsrb} &   &  $\checkmark$ & 50K+  & 43 & $15\times15$ to $250\times250$  &  Germany  \\
STS & 2011 &~\cite{larsson2011using} & $\checkmark$  &  $\checkmark$ & 3K  & 20 & $1280\times960$  &  Sweden  \\
GTSDB & 2013 &~\cite{houben2013detection} &  $\checkmark$ &  $\checkmark$ & 900   & 3 & $1360\times800$ &  Germany  \\
BelgiumTS & 2011 &~\cite{timofte2014multi} &   &  $\checkmark$ & 9K  & 62 & $11\times10$ to $562\times438$  &  Belgium  \\
LISA  &  2011  &~\cite{mogelmose2012vision} & $\checkmark$ &  $\checkmark$ &  6K  & 47 &  $640\times480$ to $1024\times522$  &  USA  \\
TT100K   &  2016  &~\cite{zhu2016traffic}  &  $\checkmark$ &  $\checkmark$&  100K  & 221 &  $2048\times2048$  &  China  \\
MTSD  &  2017  &~\cite{neuhold2017mapillary} & $\checkmark$ &  $\checkmark$ &  100K  & 313 & $920\times1080$   &  Worldwide  \\
\hline
\end{tabular}
}
\end{table*}

\section{Background}

\subsection{Adversarial attacks}

DNN predictions can be manipulated via small pixel changes in input. An adversarial attack can be performed either in a \textit{white-box} manner, where an attacker has full access to model architecture, weights, and data or in the \textit{black-box} setting, without knowledge of the target model.

One of the first and the simplest white-box adversarial attack techniques is the \textit{fast gradient sign method} (FGSM)~\cite{goodfellow2014explaining}. To obtain a perturbed image using FGSM, the sign of the gradient of the loss function with respect to the input is added to the input and multiplied by a small weighting factor. Later, iterative versions of FGSM were introduced, including the \textit{projected gradient descent} (PGD) approach~\cite{madry2017towards}. 
Other influential attack algorithms include \textit{Jacobian-based saliency maps attack} (JSMA)~\cite{papernot2017practical}, \textit{Papernot's attack}~\cite{papernot2016thelimitations}, and \textit{Carlini and Wagner} (C\&W) attack \cite{carlini2017towards}. 


In a standard setting, the attack is performed in a per-instance manner, i.e., one adversarial noise pattern is generated for each input image. However, the universal setting can be applied to make attacks more realistic, where one pattern is generated to attack the whole dataset~\cite{moosavi2017universal}. 

A further step towards realistic attacks is an \textit{adversarial patch}, where the adversarial perturbation is applied not to all input image pixels but within a specified image region~\cite{brown2017adversarial}. To perform an attack in the real world, an adversarial patch should be trained in a universal manner. Universal adversarial patch attacks have been successfully applied to various computer vision tasks, including object detection~\cite{karmon2018laVAN,pavlitskaya2022suppress,pavlitskaya2022feasibility}, semantic segmentation~\cite{nesti2022evaluating}, and steering angle prediction~\cite{pavlitskaya2020feasibility}.

Several techniques have been proposed to achieve the robustness of an adversarial patch attack under small spatial changes, camera views, and settings. For example, in the \textit{expectation over transformations} (EoT) approach by Athalye et al.~\cite{athalye2018synthesizing}, a transformation function sampled from a predefined set is applied during training, whereas possible transformations include translation, rotation, scaling, saturation, hue, and brightness changes. Eykholt et al.~\cite{eykholt2018robust} argued that the EoT approach is too constrained to model physical phenomena. Instead, they proposed the \textit{robust physical perturbations (RP2)}  method. Additionally to sampling a synthetic transformation like rotation or brightness change, training data with actual physical variability is used. This includes images taken at different distances, under different camera angles, and lighting conditions.


\subsection{Traffic sign datasets}
Traffic signs from different countries exhibit large variations, therefore a plethora of datasets for TSR has been developed (see Table \ref{tab:datasets}). Due to its age and split into traffic sign detection and classification, the German Traffic Sign Recognition and Detection Benchmark (GTSRB and GTSRD)~\cite{stallkamp2011gtsrb,houben2013detection} is usually used to evaluate TSR algorithms.  More recent approaches are often evaluated on the TsinghuaTencent 100K (TT100K) dataset~\cite{zhu2016traffic} since TSR and TSD can be combined. A profound comparison of available datasets can be found in~\cite{gray2022glare} and~\cite{wali2019vision}.

\subsection{Traffic sign detection and recognition}
Since traffic signs are characterized by specific forms and colors, early works relied predominantly on classic computer vision and image processing approaches, like SIFT~\cite{gonzalez2014text}, HOG~\cite{ellahyani2016traffic} or Hough transform~\cite{romadi2014detection}. Later, CNNs were first proposed to replace certain steps in the pipeline (e.g., feature extraction and classification) and quickly became the dominating approach in this area.

Sermanet et al.~\cite{sermanet2011traffic} were the first to surpass human results on the GTSRB by applying a multi-scale CNN (\textit{MS-CNN}) without the usage of any handcrafted features. The publicly available implementation of this architecture by Yadav\footnote{\url{https://github.com/vxy10/p2-TrafficSigns}} was often used for benchmarking in later works. Cires\c can et al.~\cite{ciresan2011committee} utilized a committee of a CNN and a multilayer perceptron with histograms of oriented gradients as input for TSR. A similar approach was taken by Jin et al.~\cite{jin2014traffic} who used an ensemble of 20 hinge loss-trained CNNs.

To go beyond the image classification task and detect traffic signs in an input image, several works tried to adapt existing generic object detectors to the TSR task. For instance, Doval et al.~\cite{doval2019traffic} utilized YOLOv3~\cite{redmon2018yolov3} for TSR on stereo camera data. Rehman et al.~\cite{rehman2022small} further optimized the architecture and training of YOLOv3  for TSR by utilizing layer pruning and a density-based anchor box selection algorithm. Another single-shot approach was suggested by Tian et al.~\cite{tian2019traffic}, who integrated a recurrent attention mechanism into the deconvolutional single shot detector~\cite{fu2017dssd} for improved usage of local context information.

Finally, Garcia et al. combined alternating convolutional and spatial transformer modules in an architecture, further denoted as \textit{CNN-ST}, outperforming state-of-the-art models on the GTSRB dataset~\cite{garcia2018deep}.

\begin{table*}[t]
\centering
\caption{Overview of adversarial attacks on traffic sign recognition models}
\label{tab:attacks-overview}
\resizebox{1.0\linewidth}{!}{
    \begin{tabular}{|l  c c | l l l l c  c  c  c  c   c  c |}
    \hline
      \textbf{Author} & \textbf{Year} & \textbf{Ref} & \textbf{Baseline}  & \textbf{Dataset} & \makecell[l]{\textbf{Attack}\\\textbf{method}} & \makecell[l]{\textbf{Attack}\\\textbf{appearance}} & \textbf{TSD} &\textbf{TSC}  &\makecell{\textbf{White-}\\\textbf{box}} & \makecell{\textbf{Black-}\\ \textbf{box}} &\makecell{\textbf{Tar-}\\ \textbf{geted}} &  \makecell{\textbf{Real-}\\ \textbf{world}} & \makecell{\textbf{Code}\\ \textbf{available}}\\ \hline

    Eykholt  et al. & 2017 &~\cite{eykholt2018robust} & \makecell[l]{LISA-CNN, GTSRB-CNN \\ (MS-CNN~\cite{sermanet2011traffic})} & LISA, GTSRB & \makecell[l]{RP2} & \makecell[l]{Black and  \\white stickers} & &  $\checkmark$ &  $\checkmark$ &  &  &  $\checkmark$ &  $\checkmark$\\ \hline
    
    Lu et al. & 2017 &~\cite{lu2017adversarial} & \makecell[l]{Faster R-CNN, \\YOLOv2} &  COCO, private   & Iterative FGSM & \makecell[l]{Perturbed sign} & $\checkmark^*$ &   &  $\checkmark$ &   &   &  $\checkmark$  &   \\\hline

    Song  et al. & 2018 &~\cite{song2018physical} & \makecell[l]{Faster-RCNN, \\YOLOv2} &COCO & \makecell[l]{RP2, modified\\for the detection task}  & \makecell[l]{Stickers or \\perturbed sign}& $\checkmark^*$ &  &  $\checkmark$ &    & &  $\checkmark$ &   \\\hline
    
    Chen et al. & 2018 &~\cite{chen2018shapeshifter} &  Faster R-CNN  &   COCO & C\&W with EoT & \makecell[l]{Perturbed sign} & $\checkmark^*$ &  &   $\checkmark$ &  & $\checkmark$  &  $\checkmark$  &  $\checkmark$ \\      \hline

    Papernot  et al. & 2017 &~\cite{papernot2017practical} & Own CNN & GTSRB & \makecell[l]{Substitute model, FGSM, \\Papernot's attack}  &\makecell[l]{Perturbed sign}&  &  $\checkmark$ &  &   $\checkmark$ &  &  $\checkmark$  & \\ \hline
    
    Sitawarin et al. & 2018 &~\cite{sitawarin2018rogue} & \makecell[l]{LISA-CNN, GTSRB-CNNN} & LISA, GTSRB & \makecell[l]{C\&W attack with EoT} &\makecell[l]{Logo, custom sign} & &  $\checkmark$ &  $\checkmark$ &   &  $\checkmark$  &  $\checkmark$ &   $\checkmark$  \\\hline
    
    Sitawarin et al. & 2018 &~\cite{sitawarin2018darts} & \makecell[l]{TSD: Hough transform\\ TSC: MS-CNN~\cite{sermanet2011traffic}} &  \makecell[l]{GTSRB, GTSDB, \\private}  &   \makecell[l]{Enhancement of~\cite{sitawarin2018rogue},\\lenticular printing attack} & \makecell[l]{Same as~\cite{sitawarin2018rogue}, signs\\looking differently\\ from various angles} & $\checkmark$ &  $\checkmark$&  $\checkmark$ &  $\checkmark$ &    $\checkmark$  &  $\checkmark$  &  $\checkmark$ \\ \hline
    
    Liu et al. & 2019 &~\cite{liu2019perceptive}  & \makecell[l]{VGG16, ResNet-34, \\MS-CNN~\cite{sermanet2011traffic}}  &   \makecell[l]{GTSRB, \\ImageNet} & PS-GAN & \makecell[l]{Scrawl-like\\stickers}& & $\checkmark$  &  $\checkmark$ & $\checkmark$ &      &  $\checkmark$  &  \\ \hline
    
    Morgulis et al. & 2019 &~\cite{morgulis2019fooling} & \makecell[l]{MS-CNN~\cite{sermanet2011traffic}, DenseNet} & GTSRB & \makecell[l]{Enhancement of \cite{sitawarin2018darts},\\improved augmentation}& \makecell[l]{Gray shadows on\\speed limit signs}&   &  $\checkmark$  &   $\checkmark$   &  $\checkmark$&      $\checkmark$    &  $\checkmark$  &  \\ \hline
    
    Li et al. & 2021 &~\cite{li2021adaptive} & \makecell[l]{CNN-ST~\cite{garcia2018deep}}
    &   GTSRB  &  \makecell[l]{Adaptive square attack, \\ SimBA-DCT \cite{guo2019simple}, \\ square attack \cite{andriushchenko2020square}} & \makecell[l]{Perturbed image}&  &  $\checkmark$  &   &  $\checkmark$  &  $\checkmark$   &   &  \\ \hline
    
    Yang et al. & 2021 &~\cite{yang2021targeted} & 3-layer CNN &  LISA, GTSRB  & \makecell[l]{Targeted attention attack} & \makecell[l]{Grayscale noises} & &  $\checkmark$ &   $\checkmark$  &    &   &  $\checkmark$ & $\checkmark$\\\hline
    
    Woitschek et al. & 2021 &~\cite{woitschek2021physical} & \makecell[l]{CNN-ST~\cite{garcia2018deep}} &   GTSRB   & \makecell[l]{FGSM, PGD, RP2, \\ SPSA, model stealing}&  \makecell[l]{Perturbed sign}  & &  $\checkmark$  &  &  $\checkmark$ &  $\checkmark$   &    &   \\\hline
    
    Jia et al. & 2021 &~\cite{jia2022fooling} & YOLOv5 &  TT100K & \makecell[l]{Cross-domain conversion} &  \makecell[l]{Perturbed sign} & $\checkmark$& $\checkmark$ &  &$\checkmark$&   $\checkmark$   &   $\checkmark$    &  \\\hline

    Ye et al. & 2021 &~\cite{ye2021patch} & \makecell[l]{VGG16, ResNet-34,\\GoogLeNet, ensemble }  & GTSRB  & \makecell[l]{PGD } & \makecell[l]{Squared patch\\in an image} &   &  $\checkmark$ &  $\checkmark$& &   &  $\checkmark$  &   \\\hline
    
    Zolfi et al. & 2021 &~\cite{zolfi2021translucent} & \makecell[l]{YOLOv2, YOLOv5,\\ Faster R-CNN}  &  \makecell[l]{LISA, MTSD,\\BDD} & \makecell[l]{Custom gradient-based\\optimization} & \makecell[l]{Translucent patch \\ on camera lens} & $\checkmark^*$  &  $\checkmark$ &  $\checkmark$& &    $\checkmark$   &  $\checkmark$  &  \\\hline
    
    Zhong et al. & 2022 &~\cite{zhong2022shadows} &  \makecell[l]{LISA-CNN, GTSRB-CNN} &  LISA, GTSRB & \makecell[l]{EoT, PSO} & \makecell[l]{Shadows} & & $\checkmark$ &    &  $\checkmark$ &    $\checkmark$   &  $\checkmark$ &  $\checkmark$\\\hline
    
    Chi et al. & 2023 &~\cite{chi2023public} &  \makecell[l]{MCDNN~\cite{ciresan2012multi}, CNN-ST~\cite{garcia2018deep}}  &      GTSRB & Public-attention attack & \makecell[l]{Perturbed sign} & &  $\checkmark$&  &    $\checkmark$  &     &   & \\\hline

    Liu et al. & 2023 &~\cite{liu2023adversarial} &  \makecell[l]{GAN}  & GTSRB, TT100K & \makecell[l]{GAN to generate\\raindrops} & \makecell[l]{Raindrops} & $\checkmark$ &  $\checkmark$&  &    $\checkmark$  &     &   & \\\hline

     Wei et al. & 2023 &~\cite{wei2023adversarial} &  YOLOv1   &  TT100K & \makecell[l]{Region-based heuristic \\differential evolution} & \makecell[l]{Perturbations of\\existing stickers} & $\checkmark$ &  $\checkmark$&  &    $\checkmark$  &     &  $\checkmark$ & $\checkmark$\\\hline
    \multicolumn{3}{l}{\textsuperscript{*}\footnotesize{within generic object detection}}
    \end{tabular}
}
\end{table*}

\begin{figure*}[t]
\centering
\resizebox{1.0\linewidth}{!}{
 \begin{tabular}{c c c c c c c}
 \includegraphics[width=0.13\textwidth]{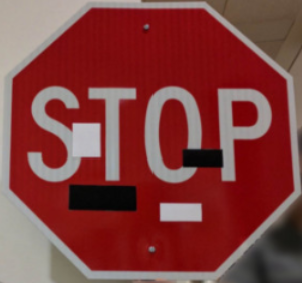} &
 \includegraphics[width=0.13\textwidth]{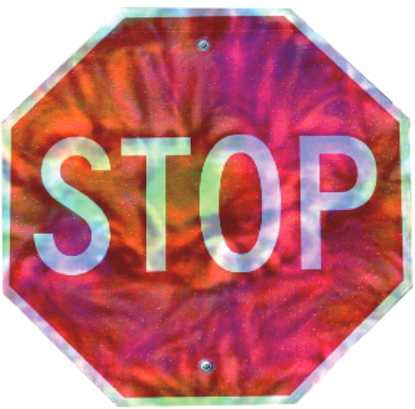} &
 \includegraphics[width=0.13\textwidth]{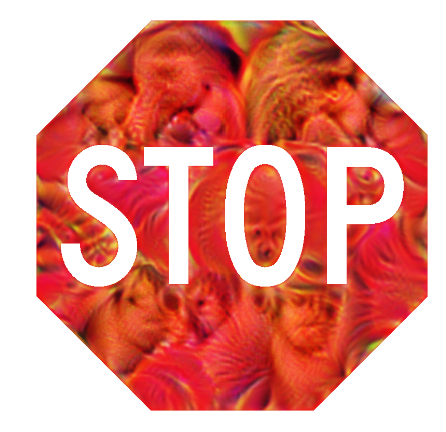} &
 \includegraphics[width=0.13\textwidth]{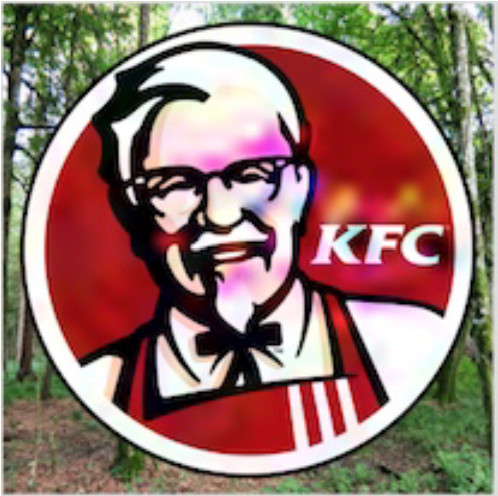} &
 \includegraphics[width=0.13\textwidth]{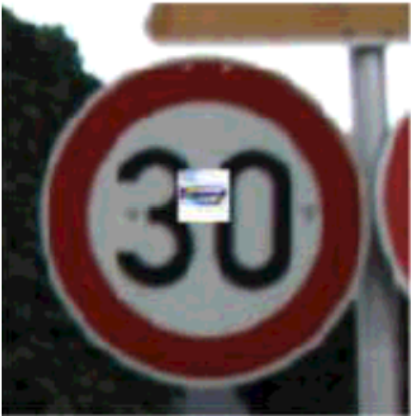} &
 \includegraphics[width=0.13\textwidth]{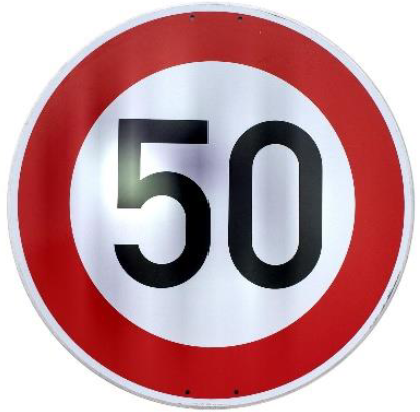} &
 \includegraphics[width=0.13\textwidth]{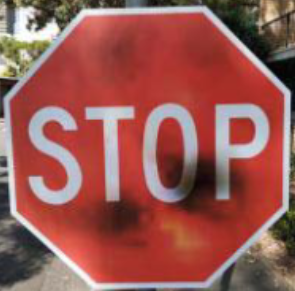} \\
 RP2: Eykholt et al.~\cite{eykholt2018robust} & Lu et al.~\cite{lu2017adversarial} & Shapeshifter: & Rogue, DARTS:  & PS-GAN~\cite{liu2019perceptive} & Morgulis et al.~\cite{morgulis2019fooling} & TAA: Yang et al.~\cite{yang2021targeted}\\
 & &  Chen et al.~\cite{chen2018shapeshifter} & Sitawarin et al.~\cite{sitawarin2018rogue,sitawarin2018darts} & & & \\
 \end{tabular}
}
    \caption{Examples of traffic signs modified with different attacks}
    \label{fig:examples}
\end{figure*}

\section{Overview of adversarial attacks on TSR}

In this section, we survey the existing adversarial attacks. Table \ref{tab:attacks-overview} provides a summarized overview, and Figure \ref{fig:examples} demonstrated examples of traffic signs modified with selected attacks.



\subsection{Early attack on TSC}

The first works on adversarial attacks against TSR appeared in 2017~\cite{papernot2017practical,eykholt2018robust,lu2017adversarial}. The seminal work by Eykholt et al.~\cite{eykholt2018robust} focused on real-world attacks against TSC models. For this, the \textit{robust physical perturbations (RP2)} approach was proposed, where a perturbation is sampled from a distribution of physically possible perturbations with the goal of maximizing the classification error. The attack was designed to resemble graffiti, usually observed on traffic signs. For this, a mask to project generated adversarial perturbations was applied to the input image. The resulting manipulation consists of a set of black and white stickers printed out and attached to the traffic sign (see Figure~\ref{fig:examples}). To account for a spectrum of printable colors, the \textit{non-printability score (NPS)} was added as a separate loss term as proposed by Sharif et al.~\cite{sharif2016accessorize}.

Furthermore, Eykholt et al. proposed a two-stage evaluation methodology for real-world attacks, including (1) stationary lab experiments and (2) field evaluation using drive-by scenarios. The authors also defined LISA-CNN and GTSRB-CNN models, which later became standard for further attack studies. LISA-CNN consists of three convolutional layers followed by a fully connected layer trained to classify images from the LISA dataset. GTSRB-CNN is based on the MS-CNN~\cite{sermanet2011traffic} implementation by Yadav, mentioned above. Models and the attack algorithm were made publicly available by the authors\footnote{\url{https://github.com/evtimovi/robust_physical_perturbations}}. 


\subsection{Attacks on the \textit{stop sign} class of generic object detection}
While the original work of Eykholt et al. focused on classifiers only, subsequent work from the same research group by Song et al.~\cite{song2018physical} also looked into detection. For this, a generic object detector YOLOv2~\cite{redmon2018yolo9000} was attacked by putting stickers on a stop sign, causing mislabeling or not detecting the traffic sign. The evaluation, however, was performed on a class \textit{stop sign} within generic object detection. Trained on the COCO dataset, YOLO can only detect stop signs. Furthermore, Song et al. expanded in this work RP2 from the classification to detection. For this, the adversarial loss was modified to cause the disappearance of the stop sign or detection of non-existent objects. Next, the set of synthetic physical constraints for RP2 was modified to include object rotation and position. And finally, the total variation~\cite{mahmood2016accesorize} was incorporated into the loss function to ensure smooth transitions between neighboring pixels. The resulting perturbation was applied to the whole stop sign area to cause object disappearance or a non-existent object creation attack. The experiments were also performed in the real world but only in stationary lab settings.

The poster attacks on stop signs performed by Eykholt et al. were critically addressed by Lu et al.~\cite{lu2017stop}. The generic detectors YOLO and Faster-RCNN were shown to be able to successfully withstand attacks proposed so far, so the authors claimed that \textit{"standard detectors aren’t (currently) fooled by physical adversarial stop signs"}. In their subsequent work, however, Lu et al. were able to successfully fool a detector themselves~\cite{lu2017adversarial}. Although the method was designed for generic object detectors, it was exemplarily evaluated on the \textit{stop sign} class. A method similar to iterative FGSM was used to generate a perturbation for the stop sign while variations in illumination intensity were incorporated. As a result of applying the generated perturbation to the stop sign, it was correctly localized in an image but classified as an object of some other class, e.g., a \textit{vase}. The authors attacked Faster R-CNN~\cite{ren2015faster} and also demonstrated that the generated perturbations transfer without modifications to YOLO2. For the real-world attacks, the authors collected their own dataset consisting of videos of stop signs from an  ego vehicle perspective. To successfully fool a model in the real world, however, large visible perturbations in the stop sign area were needed.

The \textit{Shapeshifter} attack by Chen et al.~\cite{chen2018shapeshifter} applied EOT by Athalye et al.~\cite{athalye2018synthesizing} instead of RP2 described above to ensure patch robustness in real-world settings.  This is also the first work to focus on targeted attacks on stop sign detection within the generic object detection framework. In particular, classes \textit{person} and \textit{sport ball} were selected as targets due to their similarity to stop signs in shape and size.

In summary, attacking the \textit{stop sign} class prediction of a model trained for generic object detection was addressed in multiple early works \cite{lu2017stop,lu2017adversarial,song2018physical,chen2018shapeshifter} and was extensively evaluated in real-world settings. It has paved the path for research on TSC and TSD attacks but has become less popular after that. A recent work by Zolfi et al.~\cite{zolfi2021translucent} addressed  generic object detection again and proposed a translucent patch placed directly on a camera lens.  This sticker suppressed the detection of the objects of the \textit{stop sign} class.





%

\subsection{Black-box attacks on TSR}
While most studies included the experiments with the transferability of the proposed attacks to further models, Papernot et al.~\cite{papernot2017practical} were the first to apply black-box attacks to the TSC task on GTSRB. The attacker in this setting is only able to observe the predicted labels for given inputs. The authors proposed to train a local substitute model for the attacked DNN by using its observed predictions as training data. The attack generated for the substitute DNN was shown to be able to transfer successfully to the target DNN. 

More challenging black-box attacks were the focus of recent works. Woitschek et al. studied several black-box attacks in a real-world attack setting~\cite{woitschek2021physical}. The publicly available implementation\footnote{\url{https://github.com/poojahira/
gtsrb-pytorch}} of the CNN-ST~\cite{garcia2018deep} was used for experiments. The attack algorithms included gradient approximation using \textit{simultaneous perturbation stochastic approximation} (SPSA)~\cite{spall1992multivariate} as well as \textit{model stealing}. To evaluate attack feasibility, 1000 different transformations were applied. However, no evaluation in the real world was performed.

A further score-based black-box attack method is the \textit{adaptive square attack} (ASA) proposed by Li et al. in~\cite{li2021adaptive}. It builds upon the \textit{square attacks} method by Andriuschenko et al.~\cite{andriushchenko2020square}, which uses random search, whereas, in each perturbation, square-shaped updates at random positions are sampled.   

Zhong et al.~\cite{zhong2022shadows} proposed to construct adversarial shadows by querying the target model. The attention-based attack was described by Chi et al.~\cite{chi2023public}. Wei et al.~\cite{wei2023adversarial} demonstrated a method to generate adversarial stickers against TSD in a black-box manner.

\subsection{Innocuous-looking traffic signs}
Instead of modifying existing traffic signs, Sitawarin et al.\cite{sitawarin2018rogue} proposed a method to modify innocuous signs and advertisements so that they are classified as targeted traffic signs. In this work, a traffic sign is cut out of an input image using masking and resized to reach the input size of the model. After that, a variant of the C\&W attack combined with EoT is used to generate the perturbation. An attacker can either modify an existing logo or advertisement sign or start with a blank sign to generate a custom sign. In the follow-up work~\cite{sitawarin2018darts}, the authors further enhance the proposed pipeline named \textit{DARTS} and describe the lenticular printing attack based on optical phenomena, s.t. an adversarial sign appears different under different observation angles.

Morgulis et al.~\cite{morgulis2019fooling} further extended the DARTS pipeline with improved random augmentation techniques so that a batch of new random transformations is applied to each iteration. This way, perturbations for speed limit traffic signs could be created. Printing-size adversarial signs were evaluated in the black-box mode with a TV-in-the-loop method and a drive-by experiment. 

\subsection{More realistic-looking attacks}
To further enhance the inconspicuousness of the generated stickers, Liu et al.~\cite{liu2019perceptive} used a generative adversarial network (GAN) to produce more natural-looking stickers. An attention mechanism was applied to determine areas on the traffic sign which are especially favorable for placing adversarial stickers. The \textit{targeted attention attack} (TAA) proposed by Yang et al.~\cite{yang2021targeted} generates shadow- or spot-looking perturbations on a traffic sign. The method relies on the soft attention map to find the most susceptible pixels for an attack. Zhong et al.~\cite{zhong2022shadows} also explicitly generated shadow-looking perturbations. Recently, perturbations imitating raindrops have been described by Liu et al.~\cite{liu2023adversarial}.



\section{Discussion and Conclusion}

This survey presented an overview of the existing works on adversarial attacks against traffic sign detection and classification models. In the following, we summarize our observations.

\textbf{Attacks on TSC vs. TSD}: Attacks on classification models still remain predominant in the literature. Since the seminal work by Eykholt et al.\cite{eykholt2018robust}, the evaluation stayed focused on LISA and GTSRB datasets. Attacks on TSD became popular with the introduction of GTSDB and especially TT100K datasets. Recent works \cite{wei2023adversarial,jia2022fooling,liu2023adversarial} demonstrated successful attacks using the TT100K dataset. TSC baseline models evolved from simple three-layer CNNs like LISA-CNN or multi-scale GTSRB-CNN~\cite{eykholt2018robust} to more sophisticated architectures, like a CNN with alternating convolutional and spatial transformer modules~\cite{garcia2018deep}. On the other hand, for TSD, mostly generic object detectors like YOLO were used.

\textbf{Attack appearance and physical feasibility}: Early works proposed black and white stickers~\cite{eykholt2018robust} and perturbing the whole sign area~\cite{lu2017adversarial} as an attack method. Recently, more sophisticated attacks were described, including a translucent patch~\cite{zolfi2021translucent}, shadows~\cite{yang2021targeted,zhong2022shadows}, raindrops~\cite{liu2023adversarial}, and emoji stickers~\cite{sava2022assessing}. Although attacks against TSR can easily be evaluated under realistic settings using printed traffic signs, there is a large gap between the feasibility of digital and real-world attacks. As shown by Lu et al.~\cite{lu2017adversarial}, large visible perturbations are needed to perform successful attacks in the real world. Furthermore, only several works go beyond stationary field experiments and perform drive-by field experiments. 

\textbf{Defense methods:} Attack mitigation strategies especially for the traffic sign classification and detection tasks have received significantly less attention than attacks themselves. The work by Aung et al.~\cite{aung2017building} is one of the few which evaluated adversarial training and knowledge distillation to mitigate the attacks. In this work, a CNN for TSC on the GTSRB dataset was attacked with FGSM and JSMA. Defensive distillation and adversarial training were applied by Papernot et al.~\cite{papernot2016distillation}. Recently, Zhang et al. proposed YOLOv2~\cite{redmon2018yolo9000} enhanced with adversarial patches during training \cite{zhang2022research}. Furthermore, an attempt to apply provable defenses was performed by Croce et al.~\cite{croce2019provable}.

In summary, our survey has demonstrated, how attacks on TSR have evolved from simple examples to more feasible attacks on detection and classification models. It has also shown, that research was mostly restricted to repeating baseline models and experiment settings. Furthermore, defense methods have not been studied extensively so far. We hope our overview of the existing evidence paves the way for research in this area.

\section*{Acknowledgment}

This work was supported by KASTEL Security Research Labs.

\bibliographystyle{IEEEtran}
\bibliography{references.bib}

\begin{thebibliography}{10}
\providecommand{\url}[1]{#1}
\csname url@samestyle\endcsname
\providecommand{\newblock}{\relax}
\providecommand{\bibinfo}[2]{#2}
\providecommand{\BIBentrySTDinterwordspacing}{\spaceskip=0pt\relax}
\providecommand{\BIBentryALTinterwordstretchfactor}{4}
\providecommand{\BIBentryALTinterwordspacing}{\spaceskip=\fontdimen2\font plus
\BIBentryALTinterwordstretchfactor\fontdimen3\font minus
  \fontdimen4\font\relax}
\providecommand{\BIBforeignlanguage}[2]{{%
\expandafter\ifx\csname l@#1\endcsname\relax
\typeout{** WARNING: IEEEtran.bst: No hyphenation pattern has been}%
\typeout{** loaded for the language `#1'. Using the pattern for}%
\typeout{** the default language instead.}%
\else
\language=\csname l@#1\endcsname
\fi
#2}}
\providecommand{\BIBdecl}{\relax}
\BIBdecl

\bibitem{szegedy2013intriguing}
C.~Szegedy, W.~Zaremba, I.~Sutskever, J.~Bruna, D.~Erhan, I.~Goodfellow, and
  R.~Fergus, ``{Intriguing Properties of Neural Networks},''
  \emph{International Conference on Learning Representations (ICLR)}, 2014.

\bibitem{goodfellow2014explaining}
I.~J. Goodfellow, J.~Shlens, and C.~Szegedy, ``{Explaining and Harnessing
  Adversarial Examples},'' in \emph{International Conference on Learning
  Representations (ICLR)}, 2015.

\bibitem{brown2017adversarial}
T.~B. Brown, D.~Man{\'{e}}, A.~Roy, M.~Abadi, and J.~Gilmer, ``Adversarial
  {P}atch,'' in \emph{Advances in Neural Information Processing Systems (NIPS)
  - Workshops}, 2017.

\bibitem{chen2018shapeshifter}
S.~Chen, C.~Cornelius, J.~Martin, and D.~H.~P. Chau, ``Shapeshifter: Robust
  physical adversarial attack on faster {R-CNN} object detector,'' in
  \emph{Machine Learning and Knowledge Discovery in Databases - European
  Conference, {ECML} {PKDD}}.\hskip 1em plus 0.5em minus 0.4em\relax Springer,
  2018.

\bibitem{eykholt2018robust}
K.~Eykholt, I.~Evtimov, E.~Fernandes, B.~Li, A.~Rahmati, C.~Xiao, A.~Prakash,
  T.~Kohno, and D.~Song, ``Robust physical-world attacks on deep learning
  visual classification,'' in \emph{Conference on Computer Vision and Pattern
  Recognition (CVPR)}, 2018.

\bibitem{wei2023adversarial}
X.~Wei, Y.~Guo, and J.~Yu, ``Adversarial sticker: {A} stealthy attack method in
  the physical world,'' \emph{IEEE Transactions on Pattern Analysis and Machine
  Intelligence}, 2023.

\bibitem{sermanet2011traffic}
P.~Sermanet and Y.~LeCun, ``Traffic sign recognition with multi-scale
  convolutional networks,'' in \emph{International Joint Conference on Neural
  Networks (IJCNN)}.\hskip 1em plus 0.5em minus 0.4em\relax {IEEE}, 2011.

\bibitem{garcia2018deep}
{\'{A}}.~A. Garc{\'{\i}}a, J.~A. {\'{A}}lvarez, and L.~M. Soria{-}Morillo,
  ``Deep neural network for traffic sign recognition systems: An analysis of
  spatial transformers and stochastic optimisation methods,'' \emph{Neural
  Networks}, 2018.

\bibitem{lu2017adversarial}
J.~Lu, H.~Sibai, and E.~Fabry, ``Adversarial examples that fool detectors,''
  \emph{CoRR}, vol. abs/1712.02494, 2017.

\bibitem{zhong2022shadows}
Y.~Zhong, X.~Liu, D.~Zhai, J.~Jiang, and X.~Ji, ``Shadows can be dangerous:
  Stealthy and effective physical-world adversarial attack by natural
  phenomenon,'' in \emph{Conference on Computer Vision and Pattern Recognition
  (CVPR)}.\hskip 1em plus 0.5em minus 0.4em\relax {IEEE}, 2022.

\bibitem{stallkamp2011gtsrb}
J.~Stallkamp, M.~Schlipsing, J.~Salmen, and C.~Igel, ``The german traffic sign
  recognition benchmark: {A} multi-class classification competition,'' in
  \emph{International Joint Conference on Neural Networks (IJCNN)}.\hskip 1em
  plus 0.5em minus 0.4em\relax {IEEE}, 2011.

\bibitem{larsson2011using}
F.~Larsson and M.~Felsberg, ``Using fourier descriptors and spatial models for
  traffic sign recognition,'' in \emph{Image Analysis - Scandinavian
  Conference, {SCIA}}.\hskip 1em plus 0.5em minus 0.4em\relax Springer, 2011.

\bibitem{houben2013detection}
S.~Houben, J.~Stallkamp, J.~Salmen, M.~Schlipsing, and C.~Igel, ``Detection of
  traffic signs in real-world images: The german traffic sign detection
  benchmark,'' in \emph{International Joint Conference on Neural Networks
  (IJCNN)}.\hskip 1em plus 0.5em minus 0.4em\relax {IEEE}, 2013.

\bibitem{timofte2014multi}
R.~Timofte, K.~Zimmermann, and L.~V. Gool, ``Multi-view traffic sign detection,
  recognition, and 3d localisation,'' \emph{Machine Vision and Applications},
  2014.

\bibitem{mogelmose2012vision}
A.~M{\o}gelmose, M.~M. Trivedi, and T.~B. Moeslund, ``Vision-based traffic sign
  detection and analysis for intelligent driver assistance systems:
  Perspectives and survey,'' \emph{Transactions on Intelligent Transportation
  Systems (T-ITS)}, 2012.

\bibitem{zhu2016traffic}
Z.~Zhu, D.~Liang, S.~Zhang, X.~Huang, B.~Li, and S.~Hu, ``Traffic-sign
  detection and classification in the wild,'' in \emph{Conference on Computer
  Vision and Pattern Recognition (CVPR)}.\hskip 1em plus 0.5em minus
  0.4em\relax {IEEE} Computer Society, 2016.

\bibitem{neuhold2017mapillary}
G.~Neuhold, T.~Ollmann, S.~R. Bul{\`{o}}, and P.~Kontschieder, ``The mapillary
  vistas dataset for semantic understanding of street scenes,'' in
  \emph{International Conference on Computer Vision (ICCV)}.\hskip 1em plus
  0.5em minus 0.4em\relax {IEEE} Computer Society, 2017.

\bibitem{madry2017towards}
A.~Madry, A.~Makelov, L.~Schmidt, D.~Tsipras, and A.~Vladu, ``{Towards Deep
  Learning Models Resistant to Adversarial Attacks},'' \emph{International
  Conference on Learning Representations (ICLR)}, 2018.

\bibitem{papernot2017practical}
N.~Papernot, P.~D. McDaniel, I.~J. Goodfellow, S.~Jha, Z.~B. Celik, and
  A.~Swami, ``Practical black-box attacks against machine learning,'' in
  \emph{Asia Conference on Computer and Communications Security
  (AsiaCCS)}.\hskip 1em plus 0.5em minus 0.4em\relax {ACM}, 2017.

\bibitem{papernot2016thelimitations}
N.~Papernot, P.~D. McDaniel, S.~Jha, M.~Fredrikson, Z.~B. Celik, and A.~Swami,
  ``The limitations of deep learning in adversarial settings,'' in \emph{{IEEE}
  European Symposium on Security and Privacy, EuroS{\&}P}.\hskip 1em plus 0.5em
  minus 0.4em\relax {IEEE}, 2016.

\bibitem{carlini2017towards}
N.~Carlini and D.~A. Wagner, ``Towards evaluating the robustness of neural
  networks,'' in \emph{{IEEE} Symposium on Security and Privacy, {SP}}.\hskip
  1em plus 0.5em minus 0.4em\relax {IEEE} Computer Society, 2017.

\bibitem{moosavi2017universal}
S.~Moosavi{-}Dezfooli, A.~Fawzi, O.~Fawzi, and P.~Frossard, ``Universal
  adversarial perturbations,'' in \emph{Conference on Computer Vision and
  Pattern Recognition (CVPR)}.\hskip 1em plus 0.5em minus 0.4em\relax Computer
  Vision Foundation / {IEEE}, 2017.

\bibitem{karmon2018laVAN}
D.~Karmon, D.~Zoran, and Y.~Goldberg, ``Lavan: Localized and visible
  adversarial noise,'' in \emph{International Conference on Machine Learning
  (ICML)}.\hskip 1em plus 0.5em minus 0.4em\relax {PMLR}, 2018.

\bibitem{pavlitskaya2022suppress}
S.~Pavlitskaya, J.~Hendl, S.~Kleim, L.~M{\"{u}}ller, F.~Wylczoch, and J.~M.
  Z{\"{o}}llner, ``Suppress with a patch: Revisiting universal adversarial
  patch attacks against object detection,'' in \emph{{IEEE} International
  Conference on Electrical, Computer, Communications and Mechatronics
  Engineering (ICECCME)}.\hskip 1em plus 0.5em minus 0.4em\relax {IEEE}, 2022.

\bibitem{pavlitskaya2022feasibility}
S.~Pavlitskaya, B.~Codau, and J.~M. Z{\"{o}}llner, ``Feasibility of
  inconspicuous gan-generated adversarial patches against object detection,''
  in \emph{International Joint Conference on Artificial Intelligence (IJCAI) -
  Workshops}, 2022.

\bibitem{nesti2022evaluating}
F.~Nesti, G.~Rossolini, S.~Nair, A.~Biondi, and G.~C. Buttazzo, ``Evaluating
  the robustness of semantic segmentation for autonomous driving against
  real-world adversarial patch attacks,'' in \emph{Winter Conference on
  Applications of Computer Vision (WACV)}.\hskip 1em plus 0.5em minus
  0.4em\relax {IEEE}, 2022.

\bibitem{pavlitskaya2020feasibility}
S.~Pavlitskaya, S.~{\"U}nver, and J.~M. Z{\"{o}}llner, ``Feasibility and
  suppression of adversarial patch attacks on end-to-end vehicle control,'' in
  \emph{International Conference on Intelligent Transportation Systems
  (ITSC)}.\hskip 1em plus 0.5em minus 0.4em\relax IEEE, 2020.

\bibitem{athalye2018synthesizing}
A.~Athalye, L.~Engstrom, A.~Ilyas, and K.~Kwok, ``Synthesizing robust
  adversarial examples,'' in \emph{International Conference on Machine Learning
  (ICML)}.\hskip 1em plus 0.5em minus 0.4em\relax {PMLR}, 2018.

\bibitem{gray2022glare}
N.~Gray, M.~Moraes, J.~Bian, A.~Tian, A.~Wang, H.~Xiong, and Z.~Guo, ``{GLARE:}
  {A} dataset for traffic sign detection in sun glare,'' \emph{CoRR}, vol.
  abs/2209.08716, 2022.

\bibitem{wali2019vision}
S.~B. Wali, M.~A. Abdullah, M.~A. Hannan, A.~Hussain, S.~A. Samad, P.~J. Ker,
  and M.~B. Mansor, ``Vision-based traffic sign detection and recognition
  systems: Current trends and challenges,'' \emph{Sensors}, 2019.

\bibitem{gonzalez2014text}
{\'{A}}.~Gonzalez, L.~M. Bergasa, and J.~J.~Y. Torres, ``Text detection and
  recognition on traffic panels from street-level imagery using visual
  appearance,'' \emph{Transactions on Intelligent Transportation Systems
  (T-ITS)}, 2014.

\bibitem{ellahyani2016traffic}
A.~Ellahyani, M.~E. Ansari, and I.~E. Jaafari, ``Traffic sign detection and
  recognition based on random forests,'' \emph{Applied Soft Computing}, 2016.

\bibitem{romadi2014detection}
M.~Romadi, R.~O.~H. Thami, R.~Romadi, and R.~Chiheb, ``Detection and
  recognition of road signs in a video stream based on the shape of the
  panels,'' in \emph{International Conference on Intelligent Systems
  {SITA}}.\hskip 1em plus 0.5em minus 0.4em\relax {IEEE}, 2014.

\bibitem{ciresan2011committee}
D.~C. Ciresan, U.~Meier, J.~Masci, and J.~Schmidhuber, ``A committee of neural
  networks for traffic sign classification,'' in \emph{International Joint
  Conference on Neural Networks (IJCNN)}.\hskip 1em plus 0.5em minus
  0.4em\relax {IEEE}, 2011.

\bibitem{jin2014traffic}
J.~Jin, K.~Fu, and C.~Zhang, ``Traffic sign recognition with hinge loss trained
  convolutional neural networks,'' \emph{Transactions on Intelligent
  Transportation Systems (T-ITS)}, 2014.

\bibitem{doval2019traffic}
G.~N. Doval, A.~Al-Kaff, J.~Beltr{\'a}n, F.~G. Fern{\'a}ndez, and G.~F.
  L{\'o}pez, ``Traffic sign detection and 3d localization via deep
  convolutional neural networks and stereo vision,'' in \emph{International
  Conference on Intelligent Transportation Systems (ITSC)}.\hskip 1em plus
  0.5em minus 0.4em\relax IEEE, 2019.

\bibitem{redmon2018yolov3}
J.~Redmon and A.~Farhadi, ``Yolov3: An incremental improvement,'' \emph{CoRR},
  vol. abs/1804.02767, 2018.

\bibitem{rehman2022small}
Y.~Rehman, H.~Amanullah, M.~A. Shirazi, and M.~Y. Kim, ``Small traffic sign
  detection in big images: Searching needle in a hay,'' \emph{IEEE Access},
  2022.

\bibitem{tian2019traffic}
Y.~Tian, J.~Gelernter, X.~Wang, J.~Li, and Y.~Yu, ``Traffic sign detection
  using a multi-scale recurrent attention network,'' \emph{Transactions on
  Intelligent Transportation Systems (T-ITS)}, 2019.

\bibitem{fu2017dssd}
C.~Fu, W.~Liu, A.~Ranga, A.~Tyagi, and A.~C. Berg, ``{DSSD} : Deconvolutional
  single shot detector,'' \emph{CoRR}, vol. abs/1701.06659, 2017.

\bibitem{song2018physical}
D.~Song, K.~Eykholt, I.~Evtimov, E.~Fernandes, B.~Li, A.~Rahmati,
  F.~Tram{\`{e}}r, A.~Prakash, and T.~Kohno, ``Physical adversarial examples
  for object detectors,'' in \emph{{USENIX} Workshop on Offensive
  Technologies}.\hskip 1em plus 0.5em minus 0.4em\relax {USENIX} Association,
  2018.

\bibitem{sitawarin2018rogue}
C.~Sitawarin, A.~N. Bhagoji, A.~Mosenia, P.~Mittal, and M.~Chiang, ``Rogue
  signs: Deceiving traffic sign recognition with malicious ads and logos,''
  \emph{CoRR}, vol. abs/1801.02780, 2018.

\bibitem{sitawarin2018darts}
C.~Sitawarin, A.~N. Bhagoji, A.~Mosenia, M.~Chiang, and P.~Mittal, ``{DARTS:}
  deceiving autonomous cars with toxic signs,'' \emph{CoRR}, vol.
  abs/1802.06430, 2018.

\bibitem{liu2019perceptive}
A.~Liu, X.~Liu, J.~Fan, Y.~Ma, A.~Zhang, H.~Xie, and D.~Tao,
  ``Perceptual-sensitive {GAN} for generating adversarial patches,'' in
  \emph{AAAI Conference on Artificial Intelligence}.\hskip 1em plus 0.5em minus
  0.4em\relax {AAAI} Press, 2019.

\bibitem{morgulis2019fooling}
N.~Morgulis, A.~Kreines, S.~Mendelowitz, and Y.~Weisglass, ``Fooling a real car
  with adversarial traffic signs,'' \emph{CoRR}, vol. abs/1907.00374, 2019.

\bibitem{li2021adaptive}
Y.~Li, X.~Xu, J.~Xiao, S.~Li, and H.~T. Shen, ``Adaptive square attack: Fooling
  autonomous cars with adversarial traffic signs,'' \emph{IEEE Internet of
  Things Journal}, 2021.

\bibitem{guo2019simple}
C.~Guo, J.~R. Gardner, Y.~You, A.~G. Wilson, and K.~Q. Weinberger, ``Simple
  black-box adversarial attacks,'' in \emph{International Conference on Machine
  Learning (ICML)}.\hskip 1em plus 0.5em minus 0.4em\relax {PMLR}, 2019.

\bibitem{andriushchenko2020square}
M.~Andriushchenko, F.~Croce, N.~Flammarion, and M.~Hein, ``Square attack: {A}
  query-efficient black-box adversarial attack via random search,'' in
  \emph{European Conference on Computer Vision (ECCV)}.\hskip 1em plus 0.5em
  minus 0.4em\relax Springer, 2020.

\bibitem{yang2021targeted}
X.~Yang, W.~Liu, S.~Zhang, W.~Liu, and D.~Tao, ``Targeted attention attack on
  deep learning models in road sign recognition,'' \emph{IEEE Internet of
  Things Journal}, 2021.

\bibitem{woitschek2021physical}
F.~Woitschek and G.~Schneider, ``Physical adversarial attacks on deep neural
  networks for traffic sign recognition: {A} feasibility study,'' in
  \emph{Intelligent Vehicles Symposium (IV)}.\hskip 1em plus 0.5em minus
  0.4em\relax {IEEE}, 2021.

\bibitem{jia2022fooling}
W.~Jia, Z.~Lu, H.~Zhang, Z.~Liu, J.~Wang, and G.~Qu, ``Fooling the eyes of
  autonomous vehicles: Robust physical adversarial examples against traffic
  sign recognition systems,'' in \emph{Annual Network and Distributed System
  Security Symposium {NDSS}}.\hskip 1em plus 0.5em minus 0.4em\relax The
  Internet Society, 2022.

\bibitem{ye2021patch}
B.~Ye, H.~Yin, J.~Yan, and W.~Ge, ``Patch-based attack on traffic sign
  recognition,'' in \emph{International Conference on Intelligent
  Transportation Systems (ITSC)}.\hskip 1em plus 0.5em minus 0.4em\relax
  {IEEE}, 2021.

\bibitem{zolfi2021translucent}
A.~Zolfi, M.~Kravchik, Y.~Elovici, and A.~Shabtai, ``The translucent patch: {A}
  physical and universal attack on object detectors,'' in \emph{Conference on
  Computer Vision and Pattern Recognition (CVPR)}, 2021.

\bibitem{chi2023public}
L.~Chi, M.~Msahli, G.~Memmi, and H.~Qiu, ``Public-attention-based adversarial
  attack on traffic sign recognition,'' in \emph{{IEEE} Consumer Communications
  {\&} Networking Conference {CCNC}}.\hskip 1em plus 0.5em minus 0.4em\relax
  {IEEE}, 2023.

\bibitem{ciresan2012multi}
D.~C. Ciresan, U.~Meier, J.~Masci, and J.~Schmidhuber, ``Multi-column deep
  neural network for traffic sign classification,'' \emph{Neural Networks},
  2012.

\bibitem{liu2023adversarial}
J.~Liu, B.~Lu, M.~Xiong, T.~Zhang, and H.~Xiong, ``Adversarial attack with
  raindrops,'' \emph{CoRR}, 2023.

\bibitem{sharif2016accessorize}
M.~Sharif, S.~Bhagavatula, L.~Bauer, and M.~K. Reiter, ``Accessorize to a
  crime: Real and stealthy attacks on state-of-the-art face recognition,'' in
  \emph{{ACM} {SIGSAC} Conference on Computer and Communications
  Security}.\hskip 1em plus 0.5em minus 0.4em\relax {ACM}, 2016.

\bibitem{redmon2018yolo9000}
J.~Redmon and A.~Farhadi, ``{YOLO9000:} better, faster, stronger,'' in
  \emph{Conference on Computer Vision and Pattern Recognition (CVPR)}.\hskip
  1em plus 0.5em minus 0.4em\relax {IEEE} Computer Society, 2017.

\bibitem{mahmood2016accesorize}
M.~Sharif, S.~Bhagavatula, L.~Bauer, and M.~K. Reiter, ``Accessorize to a
  crime: {R}eal and stealthy attacks on state-of-the-art face recognition,'' in
  \emph{Conference on Computer and Communications Security (CCS)}.\hskip 1em
  plus 0.5em minus 0.4em\relax {ACM}, 2016.

\bibitem{lu2017stop}
J.~Lu, H.~Sibai, E.~Fabry, and D.~A. Forsyth, ``Standard detectors aren't
  (currently) fooled by physical adversarial stop signs,'' \emph{CoRR}, vol.
  abs/1710.03337, 2017.

\bibitem{ren2015faster}
S.~Ren, K.~He, R.~B. Girshick, and J.~Sun, ``Faster {R-CNN:} towards real-time
  object detection with region proposal networks,'' in \emph{Advances in Neural
  Information Processing Systems (NIPS)}, 2015.

\bibitem{spall1992multivariate}
J.~C. Spall, ``Multivariate stochastic approximation using a simultaneous
  perturbation gradient approximation,'' \emph{IEEE Transactions on Automatic
  Control}, 1992.

\bibitem{sava2022assessing}
P.~A. Sava, J.~Schulze, P.~Sperl, and K.~B{\"{o}}ttinger, ``Assessing the
  impact of transformations on physical adversarial attacks,'' in \emph{{ACM}
  Workshop on Artificial Intelligence and Security}.\hskip 1em plus 0.5em minus
  0.4em\relax {ACM}, 2022.

\bibitem{aung2017building}
A.~M. Aung, Y.~Fadila, R.~Gondokaryono, and L.~Gonzalez, ``Building robust deep
  neural networks for road sign detection,'' \emph{CoRR}, vol. abs/1712.09327,
  2017.

\bibitem{papernot2016distillation}
N.~Papernot, P.~D. McDaniel, X.~Wu, S.~Jha, and A.~Swami, ``Distillation as a
  defense to adversarial perturbations against deep neural networks,'' in
  \emph{{IEEE} Symposium on Security and Privacy}.\hskip 1em plus 0.5em minus
  0.4em\relax {IEEE} Computer Society, 2016.

\bibitem{zhang2022research}
Y.~Zhang, J.~Cui, and M.~Liu, ``Research on adversarial patch attack defense
  method for traffic sign detection,'' in \emph{Cyber Security - China Annual
  Conference, {CNCERT}}, W.~Lu, Y.~Z. andf Weiping~Wen, H.~Yan, and C.~Li,
  Eds., 2022.

\bibitem{croce2019provable}
F.~Croce, M.~Andriushchenko, and M.~Hein, ``Provable robustness of relu
  networks via maximization of linear regions,'' in \emph{International
  Conference on Artificial Intelligence and Statistics (AISTATS)}.\hskip 1em
  plus 0.5em minus 0.4em\relax {PMLR}, 2019.

\end{thebibliography}

\end{document}